\documentclass[10pt,twocolumn,letterpaper]{article}

\usepackage{cvpr}      

\usepackage{stfloats}

\usepackage[dvipsnames]{xcolor}

\definecolor{cvprblue}{rgb}{0.21,0.49,0.74}
\usepackage[pagebackref,breaklinks,colorlinks,citecolor=cvprblue]{hyperref}

\def\confName{CVPR}
\def\confYear{2024}

\title{Whole-Herd Elephant Pose Estimation from Drone Data for Collective Behavior Analysis}
\author{Brody McNutt\thanks{Form Bio, Austin, TX, USA}, 
    Libby Zhang\thanks{Colossal Biosciences, Dallas, TX, USA},  
    Angus Carey-Douglas\thanks{Save the Elephants, Nairobi, Kenya},   
    Fritz Vollrath\footnotemark[3] \thanks{University of Oxford, Oxford,  United Kingdom}, 
    Frank Pope \footnotemark[3], 
    Leandra Brickson \footnotemark[2] \thanks{publications@colossal.com}
 }

\begin{document}
\maketitle
\begin{abstract}

This research represents a pioneering application of automated pose estimation from drone data to study elephant behavior in the wild, utilizing video footage captured from Samburu National Reserve, Kenya. The study evaluates two pose estimation workflows: DeepLabCut, known for its application in laboratory settings and emerging wildlife fieldwork, and YOLO-NAS-Pose, a newly released pose estimation model not previously applied to wildlife behavioral studies. These models are trained to analyze elephant herd behavior, focusing on low-resolution ($\sim$50 pixels) subjects to detect key points such as the head, spine, and ears of multiple elephants within a frame. Both workflows demonstrated acceptable quality of pose estimation on the test set, facilitating the automated detection of basic behaviors crucial for studying elephant herd dynamics. For the metrics selected for pose estimation evaluation on the test set\textemdash root mean square error (RMSE), percentage of correct keypoints (PCK), and object keypoint similarity (OKS)\textemdash the YOLO-NAS-Pose workflow outperformed DeepLabCut. Additionally, YOLO-NAS-Pose exceeded DeepLabCut in object detection evaluation. This approach introduces a novel method for wildlife behavioral research, including the burgeoning field of wildlife drone monitoring, with significant implications for wildlife conservation.
\end{abstract}    
\section{Introduction}
\label{sec:intro}
More nuanced and precise understanding of elephant behavior is crucial for developing effective conservation strategies in the face of multiplying threats, such as rapid climate change and loss of habitat and migratory corridors. 
African savanna elephants (\textit{Loxodonta africana}) live in flexible fission-fusion societies that result in sophisticated social interactions and decision-making at different organization levels; thus elephant behavior is best understood concurrently at both the individual and group level \cite{Archie2006}.
Direct field observation is the established approach to studying elephant behavior at the spatial and temporal resolution required to gain insight into these types of sophisticated interactions.
A significant disadvantage, however, is the limited field-of-view of a single observer and the practical challenges of recording simultaneously behaviors from multiple animals.

Aerial-based video recording platforms are emerging as a promising approach to capturing multi-animal behavior in open terrain over greater field-of-views and spatial ranges than previously possible. 
For example, Koger et al. released a comprehensive software package with individualized detecting, tracking, and pose estimation modules \cite{Koger2023-ro}.
The emergence of aerial-based video recording platforms has been enabled by continued progress in unmanned aerial vehicle technology and in computer vision. The latter was significantly advanced by the deep learning revolution, allowing the propagation of information-dense raw data throughout all modules of the system. Other important advantages of these end-to-end deep learning solutions included simplifications in piping and parameter tuning.
With this revolution, however, also came the reports of instances in which the state-of-the-art methods could not generalize out-of-the-box to other domains, as they were purported to.
This was particularly illustrated in fields such as computer-vision-based animal pose estimation \cite{Mathis2018-jg} or animal detection \cite{DBLP:journals/corr/abs-1907-06772, Beery_Efficient_Pipeline_for}. In particular, the performance gap was due to differences such as labeled dataset sizes and challenging visual discrimination conditions that were overcome by strategies such as fine-tuning on animal-specific data \cite{Mathis2020-ed}.

In this paper, we revisit this question of modular, composite solutions, such as the one provided by Koger et al., versus end-to-end solutions given the objective of extracting multi-animal pose estimates from aerial video recordings.
We note that this task differs from overhead video recordings that might be found in laboratory settings due to the increased background complexity and variability and significantly smaller size of the subjects (8-70 px in our dataset).

This paper details the methods for adapting this data for use by DeepLabCut \cite{Mathis2018-jg} and explores the viability of YOLO-NAS-Pose \cite{terven2023comprehensive}, the former being a common tool used in behavioral neuroscience experiments, primarily in lab settings, and the latter being a state-of-the-art general-use pose estimation model, not traditionally used for behavioral research. DeepLabCut, often used in behavioral neuroscience within lab environments, however, has seen some application in wildlife studies. On the other hand, YOLO-NAS-Pose offers a streamlined workflow that requires minimal preprocessing for data formatting and benefits from rapid execution times. 
We compare the performance of a composite workflow (“DeepLabCut Workflow”) to an end-to-end pose workflow (“YOLO-NAS-Pose Workflow”) on accurately extracting pose estimates across multiple animals from aerial video recordings.

\section{Methods}
\subsection{Dataset} 

This research employs drone technology equipped with a wide-angle camera to observe a herd of elephants, ensuring visibility of all herd members in a single frame. Drone data collection introduces specific challenges. The Save the Elephants field team aimed to optimize data quality while minimally impacting the elephants to capture authentic behavior, noting from previous studies that drones can trigger varying responses from elephants \cite{bennitt2019terrestrial, hartmann2021first, mesquita2022terrestrial}. While data with higher resolution would have been advantageous, using multiple drones could have altered the elephants' natural behaviors. To mitigate this, the drone was operated at the maximum allowable height in Kenya (400ft).
\begin{figure*}
    \centering
    \begin{minipage}{0.64\linewidth}
        \includegraphics[width=\linewidth]{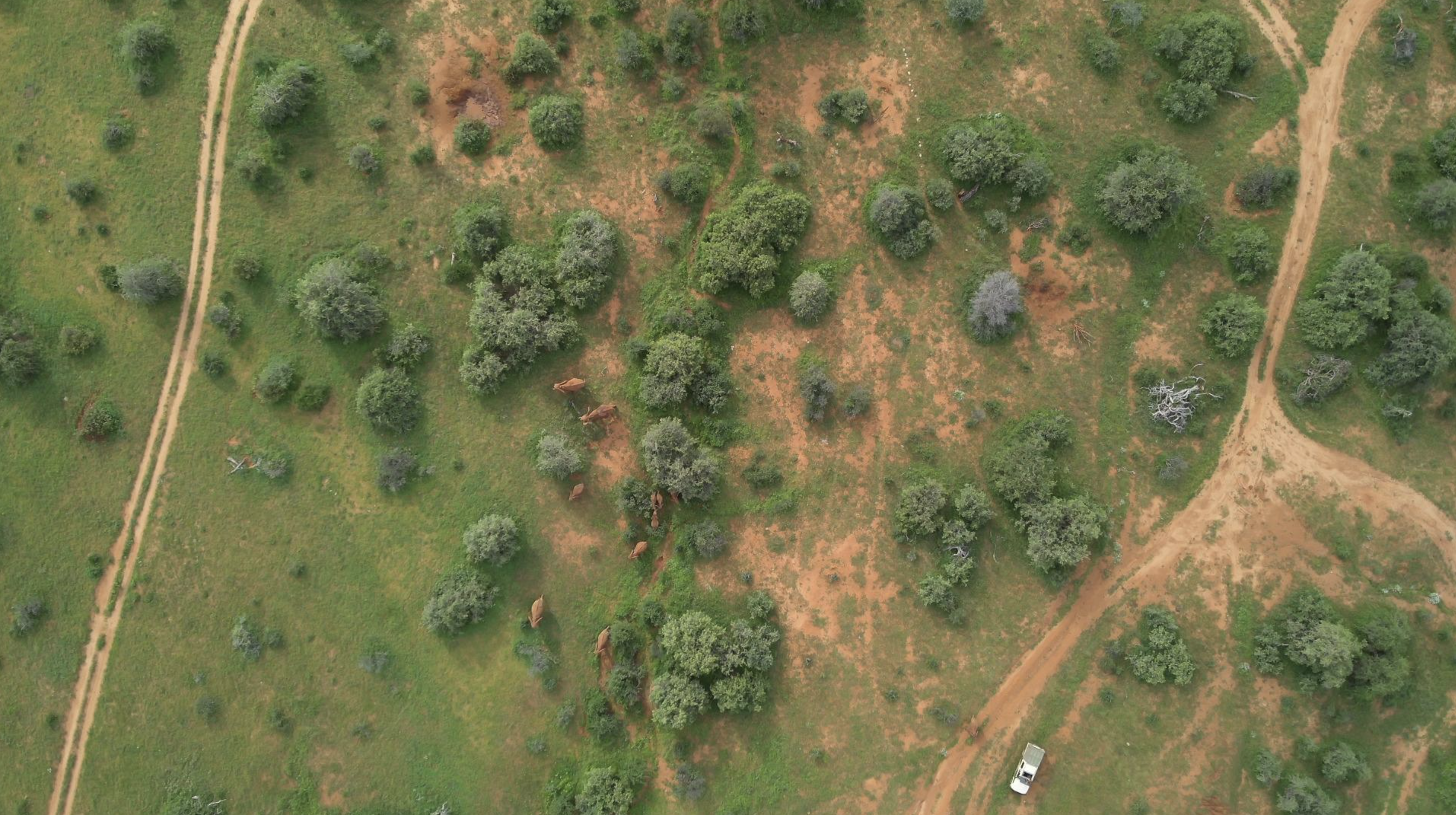}
        \caption{Example frame from captured drone footage from Save the Elephants in Samburu National Reserve, Kenya. The resolution has been greatly reduced for this manuscript.}
        \label{fig:drone}
    \end{minipage}
    \hfill
    \begin{minipage}{0.35\linewidth}
        \includegraphics[width=\linewidth]{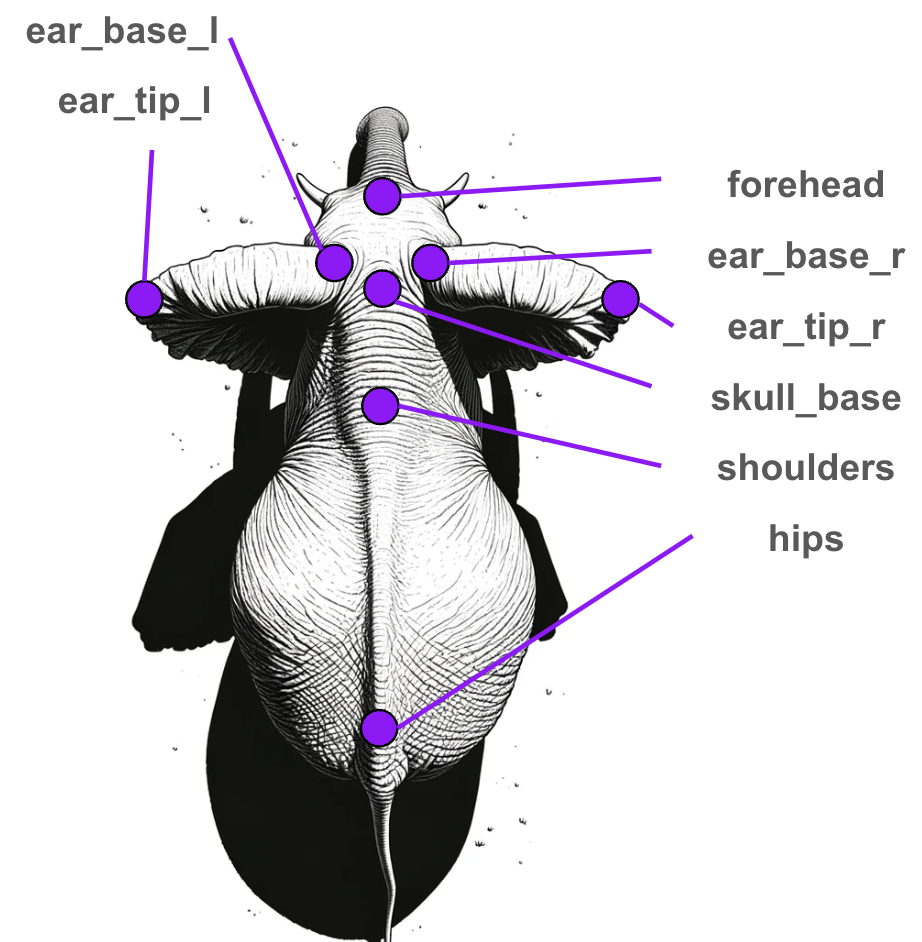}
        \caption{Keypoints of desired pose estimation in order to achieve automated detection for ear flapping, head orientation.}
        \label{fig:keypoints}
    \end{minipage}
\end{figure*}
The drone captured footage at 29 fps with a 3840x2160 resolution on a stabilizing gimbal platform. During recording, the drone is positioned stationary, overhead at a set height throughout the study to ensure a uniform viewing angle. At the drone's operating altitude for this study, calves are represented from trunk to tail by about 8 pixels, and adults by up to 70 pixels in the video footage. Figure \ref{fig:drone} showcases a sample frame from the drone footage. 

The study focused on identifying keypoints relevant to social behaviors, such as head orientation and ear flapping. Therefore, the 8 keypoints shown in Figure \ref{fig:keypoints} were chosen as our targets for pose estimation.

This dataset consists of 23 videos, each approximately 5 minutes in length. Overhead frames were selected from these videos, resulting in a total of 133 frames containing 1308 elephants. A manually annotated training dataset was created from these frames, including bounding boxes and the keypoints defined in Figure \ref{fig:keypoints}. During annotation, when the ears were not discernible on especially small calves, only spine keypoints were annotated, and the ears were labeled as “occluded”. 

The labeled dataset was divided into a 90-10-10 train-validation-test split. For this data split, the test set comprised four entirely set-aside videos, ensuring that no frames in the test set originated from the same videos as those in the training and validation sets. In contrast, while the training and validation images were unique, they could still come from the same videos.

\subsubsection{Preprocessing} Before entering either workflow, the data was preprocessed to meet the YOLOv5 model's requirements for object size \cite{Leorna2022}. Labeled video frames were tiled to 800x800 pixels, with a 33\% overlap in window stride, to ensure a proper object size for the elephants within the frame. Pose estimation was then applied to the data using the following two workflows.

\subsection{DeepLabCut Workflow}
\begin{figure*}[ht!]
    \centering
    \includegraphics[width=\textwidth]{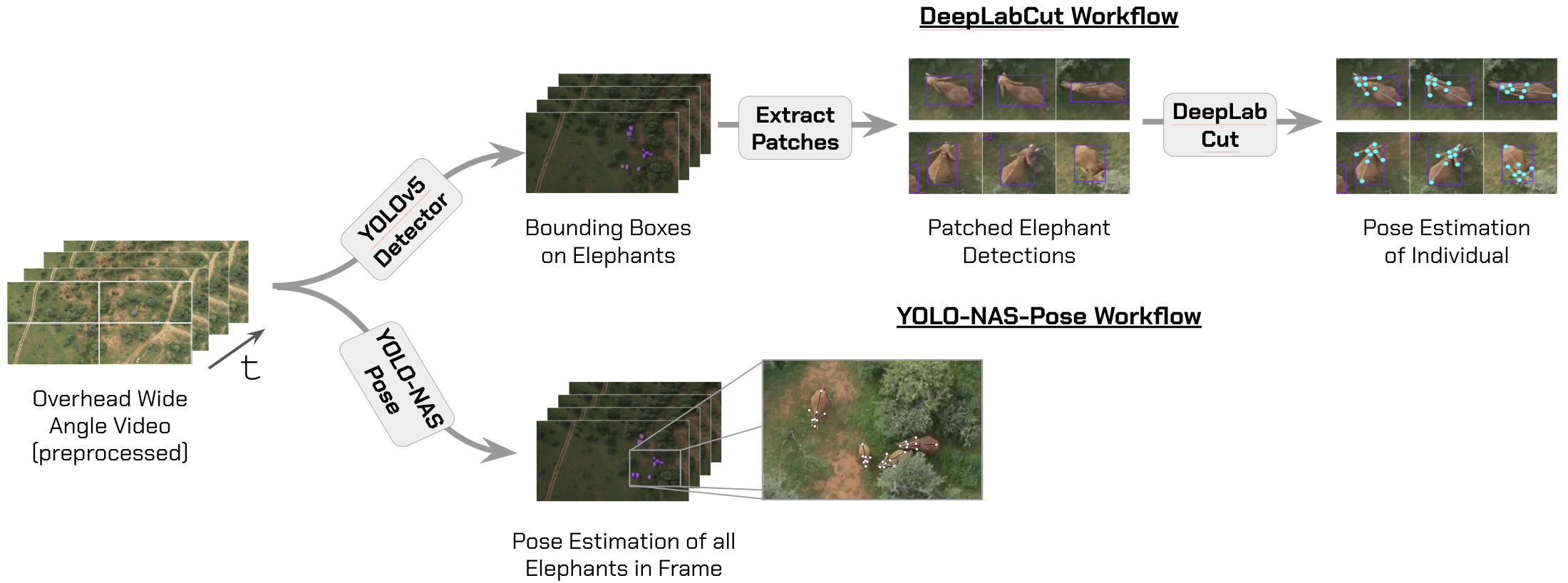}
    \caption{Workflow details for both methods investigated}
    \label{fig:my_label}
\end{figure*}

\subsubsection{Elephant Detector} 
Initially, a YOLOv5 model \cite{Jocher2022} and a MegaDetector \cite{DBLP:journals/corr/abs-1907-06772, Beery_Efficient_Pipeline_for} pretrained model were fine-tuned on the dataset defined in the previous section. The models were trained to generate bounding boxes for elephants within a given frame. 

Once bounding boxes were predicted on a frame, square images were extracted, centered on the detected bounding box, with the dimension determined by adding a 20\% margin to the largest dimension of the bounding box. These patches were then resized to 100x100 pixels. This format was used to train DeepLabCut, providing centered, large images of the animals to mitigate any unwanted effects from inconsistent backgrounds in the images.

\subsubsection{DeepLabCut}
To train DeepLabCut, the pose dataset defined in the dataset section was used to train a DeepLabCut Model. The dataset was converted to the DLC training format, and the model was trained for 800k iterations until loss converged. 

\subsection{YOLO-NAS-Pose workflow}
To train the YOLO-NAS-Pose network, the same dataset used for training the detector and DeepLabCut workflow was utilized, with manually annotated poses added. The model was then trained to provide bounding boxes and poses across the entire image. 

\subsection{Evaluation} 
The dedicated set-aside test set was used to evaluate both workflows. The bounding box accuracy for both the YOLOv5 detector and YOLO-NAS-Pose was evaluated using mean Average Precision (mAP) \cite{everingham2010pascal}. Pose estimation for both workflows was evaluated using root mean square error (RMSE), percentage of correct keypoints (PCK) \cite{Gu2020}, and object keypoint similarity (OKS) \cite{Ruggero2017}. However, to properly compare the two methods, since DeepLabCut can only perform pose estimation on extracted bounding boxes, only the bounding boxes correctly detected in the YOLO-NAS-Pose workflow were selected for pose estimation evaluation. 

To identify correctly detected objects, the bounding boxes output by YOLO-NAS-Pose were filtered using non-maximum suppression (NMS) with a maximum overlap threshold of 0.5. These de-duplicated bounding boxes were then sorted by confidence score and compared to the ground truth annotations to calculate Intersection over Union (IoU). Each predicted bounding box that shared an IoU greater than or equal to 0.5 with a ground truth bounding box was considered a candidate match. In instances where multiple predictions overlapped with the same ground truth bounding box, the prediction with the highest confidence score was selected.

\subsubsection{Video Tracker for Visualization} Although continuous video is not necessary for training or quantitatively evaluating pose estimation performance, having continuous video of an individual significantly aids in qualitative assessment. Once individuals were detected in each frame, DeepSORT \cite{Wojke2017simple, Wojke2018deep} was employed to generate patched video segments of each detected elephant. This method identifies continuous objects within the video by comparing patch locations, image embeddings, and the momentum of the objects, resulting in a sequence of bounding boxes for each individual. Due to the low resolution of some individuals, those with bounding boxes smaller than 50 pixels were excluded from this evaluation, prioritizing the analysis of adult elephant behavior. After processing, a total of 25 videos were extracted from the original source videos of the train, validation and test sets to evaluate the pose estimation on video data.  
 
\section{Results}

During the initial workflow where the YOLOv5 detector was trained, it was observed that utilizing the standard pretrained weights of YOLOv5 yielded better results compared to beginning with the megadetector weights. Mean average precision metrics for bounding box detectors are shown in Table \ref{tab:bb_metrics}
\begin{table}[ht]
    \centering
    \begin{tabular}{ccc} 
        \hline
           & YOLOv5 & YOLO-NAS-Pose \\ 
        \hline
        mAP@0.3:0.05:0.95 & 0.46 & 0.65 \\ 
        \hline
        mAP@0.5 & 0.65 & 0.81 \\ 
        \hline
    \end{tabular}
    \caption{Bounding box models test set performance: mAP@0.3:0.05:0.9 is mean Average Precision over IoU thresholds ranging from 0.3 to 0.95 with a step size of 0.05, mAP@0.5 is mean Average Precision at an IoU threshold of 0.5}
    \label{tab:bb_metrics}
\end{table}
\begin{table*}[ht]
    \centering
    \begin{tabular}{cccccccccc} 
        \hline
        & \multicolumn{3}{c}{DeepLabCut} & \multicolumn{3}{c}{YOLO-NAS-Pose} \\
        \hline
        & RMSE & PCK & OKS & RMSE & PCK & OKS \\ 
        \hline
        forehead & 6.7 & 44.4 & 0.72 & 20.12 & 0.0 & 0.02 \\    
        ear\_base\_l & 5.3 & 52.6 & 0.74 & 3.11 & 68.7 & 0.84 \\    
        ear\_base\_r & 6.7 & 49.6 & 0.73 & 2.34 & 63.9 & 0.85 \\    
        skull\_base & 5.6 & 53.4 & 0.76 & 0.08 & 64.5 & 0.82 \\    
        shoulders & 3.9 & 49.6 & 0.76 & 0.16 & 40.4 & 0.72 \\    
        hips & 9.3 & 23.3 & 0.53 & 3.22 & 67.5 & 0.82 \\    
        ear\_tip\_l & 5.9 & 32.3 & 0.60 & 3.50 & 47.0 & 0.75 \\    
        ear\_tip\_r & 6.4 & 21.1 & 0.54 & 3.36 & 53.6 & 0.75 \\    
        Average & \textbf{6.3} & \textbf{40.8} & \textbf{0.67} & \textbf{5.32} & \textbf{50.7} & \textbf{0.70} \\ 
        \hline
    \end{tabular}
    \caption{Performance metrics of DeepLabCut and YOLO-NAS-Pose models on the set-aside test set.}
    \label{tab:pose_metrics}
\end{table*}
The evaluation metrics described in the methods section were calculated on the set-aside test set, and the results for each keypoint, along with the average across all keypoints, are presented in Table \ref{tab:pose_metrics}.

\begin{figure}[ht]
    \centering
    \includegraphics[width=0.9\columnwidth]{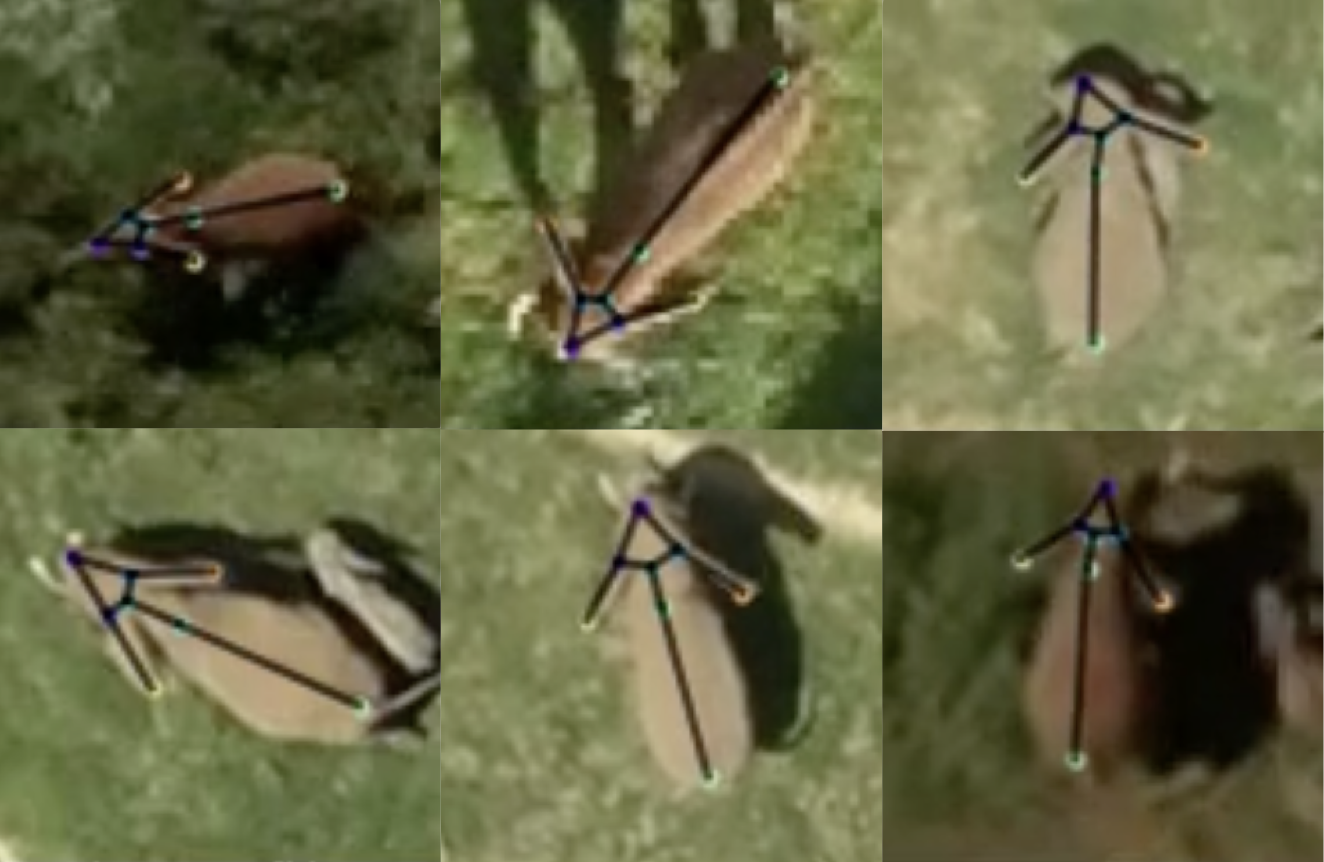}
    \caption{Example of pose estimations generated by DeepLabCut on patches extracted from the YOLOv5 detector.}
    \label{fig:dlc}
\end{figure}

Figure \ref{fig:dlc} illustrates the results of DeepLabCut applied to the extracted patch frames. The supplementary materials include training and validation set videos from the video tracker with the pose estimation overlaid. These materials showcase examples where DeepLabCut performs well, as well as select instances where the results are suboptimal. In these examples, while spinal alignment is consistently maintained, inaccuracies are noted in the ear tip detection, particularly during swift movements or uncommon poses.

\begin{figure}[ht]
    \centering
    \includegraphics[width=0.9\columnwidth]{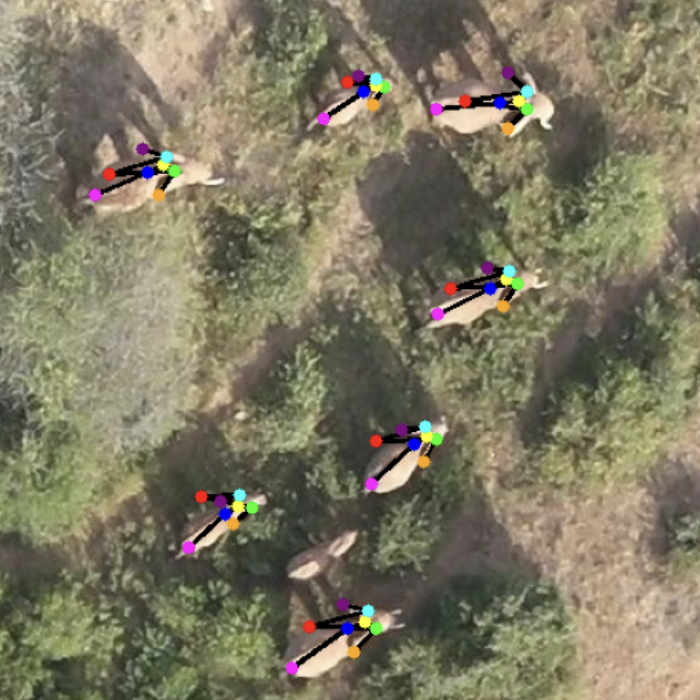}
    \caption{Example of a test set image with pose estimations YOLO-NAS-Pose overlaid. Though there is decent performance, with only one false-positive calf, the “forehead” keypoint is consistently off for all detected elephants.}
    \label{fig:ynp}
\end{figure}

Qualitative results for YOLO-NAS-Pose for a single video frame are depicted in Figure \ref{fig:ynp}. Overall, the model correctly labels keypoints, only missing one calf in this example. However, the “forehead” keypoint is consistently mispositioned behind the head.

\section{Discussion}

This research represents a pioneering application of automated pose estimation to elephant video drone data in wildlife settings. The results provide valuable insights and opportunities for future improvements in wildlife behavioral monitoring. 

When examining the metrics in Table \ref{tab:pose_metrics}, both models demonstrate reasonable performance in pose estimation on the test dataset. YOLO-NAS-Pose performed well, though not perfectly, in both elephant detection and pose estimation across all metrics, establishing it as a promising tool for wildlife behavioral studies. However, while the results are promising, the current metrics do not yet achieve the desired level of accuracy for a fully automated workflow, indicating that further development and refinement are necessary. 

It is important to note the discrepancies in keypoint accuracy within the metrics. For DeepLabCut, the accuracy of both ear tip detections was slightly lower, which was expected due to their wide range of motion relative to other keypoints and the lowest confidence during manual annotation. However, the hips, surprisingly, had the worst keypoint accuracy for DeepLabCut. This could be attributed to the hips being the most isolated keypoint, with fewer adjacent reference points for accurate positioning. This poor performance is unexpected, particularly since the hips were one of the highest-performing keypoints for YOLO-NAS-Pose. Conversely, YOLO-NAS-Pose struggled the most with the “forehead” keypoint, an area where DeepLabCut does not experience issues. One potential reason could be the difficulty in accurately labeling the “forehead”, especially when the trunk is extended, making it challenging to locate the front of the face. Future investigations will explore the causes of these discrepancies. 

Qualitatively, from watching the tracking videos applied to the full videos which were the source of the train and validation sets, DeepLabCut performed quite well, but occasionally failed to track the elephants' ears, often defaulting to a “neutral” ear posture in uncertain cases. This issue was particularly prevalent for smaller elephants.

Another noteworthy aspect to consider is the comparison between full-frame pose estimation of multiple elephants and pose estimation of an individual in an extracted patch. These approaches offer distinct advantages. Full-frame pose estimation simplifies the workflow, making it an attractive solution for automated processes. However, segmenting out individuals first provides several benefits for training a more robust network. For example, by filtering for only large elephants during training, one can avoid the challenges of training on smaller calves whose resolution may be insufficient for accurate labeling.

Moreover, individual labels allow for better balancing of the training dataset, ensuring an even distribution of poses. This technique is crucial for training pose estimators effectively. In contrast, a random sampling of data tends to result in a dataset dominated by neutral postures, limiting the diversity of the training set. 

While DeepLabCut did not outperform YOLO-NAS-Pose in this task, there are scenarios where it can be highly useful. The supplementary materials highlight an initial experiment, not detailed in this work, demonstrating that DeepLabCut can yield satisfactory results even with very small training datasets ($\sim$100 frames). If the researcher's objective is to label a few frames in a video and subsequently obtain poses for the entire video, DeepLabCut proves to be a powerful option. This capability makes it particularly valuable for projects with limited annotated data, where rapid and efficient pose estimation is required.

Looking ahead, for low-resolution pose estimation, the challenge for detecting more complex keypoints for more detailed behavior analysis lies in detecting specific keypoints by examining video sequence changes. The difficulty of identifying an elephant's ear position in a single frame highlights the limitations of current frame-by-frame pose estimation methods, which do not consider inter-frame continuity. Investigating frame-to-frame analysis methods, such as optical flow or recurrent neural networks, could offer a means to further enhance pose estimation accuracy by ensuring consistency in detected movements across video frames.

\section{Conclusion}

This research represents a substantial advancement in integrating automated behavior analysis methods into wildlife research by comparing different pose estimation techniques. It paves the way for more sophisticated studies of wildlife behaviors in their natural habitats, involving multiple individuals in extensive scenes. The findings indicate that YOLO-NAS-Pose is a feasible and attractive option for pose estimation, offering a straightforward workflow and superior performance metrics. However, further development and refinement are necessary. The implications of this work extend beyond the study of elephant behaviors, providing valuable insights for the future development of drone-based wildlife behavior studies across various species and ecosystems.

{
    \small
    \bibliographystyle{ieeenat_fullname}
    \bibliography{main}
}


\end{document}